\pdfoutput=1

\documentclass[11pt]{article}
\usepackage{makecell}
\usepackage{array}
\usepackage{stfloats}

\usepackage[preprint]{acl}

\usepackage{times}
\usepackage{latexsym}
\usepackage{enumitem}

\usepackage[T1]{fontenc}

\usepackage[utf8]{inputenc}

\usepackage{microtype}

\usepackage{inconsolata}

\usepackage{graphicx}

\usepackage{linguex}
\title{Un-considering Contextual Information: \\
Assessing LLMs’ Understanding of Indexical Elements}

\author{
   Metehan Oğuz\quad 
   Yavuz Bakman\quad  Duygu Nur Yaldiz \\
University of Southern California \\
  \texttt{\{moguz, ybakman, yaldiz\}@usc.edu}}

\begin{document}
\maketitle
\begin{abstract}
Large Language Models (LLMs) have demonstrated impressive performances in tasks related to coreference resolution. However, previous studies mostly assessed LLM performance on coreference resolution with nouns and third person pronouns. This study evaluates LLM performance on coreference resolution with indexicals like \textit{I, you, here} and \textit{tomorrow}, which come with unique challenges due to their linguistic properties. We present the first study examining how LLMs interpret indexicals in English, releasing the English Indexical Dataset with 1600 multiple-choice questions. We evaluate pioneering LLMs, including GPT-4o, Claude 3.5 Sonnet, Gemini 1.5 Pro, and DeepSeek V3. Our results reveal that LLMs exhibit an impressive performance with some indexicals (\textit{I}), while struggling with others (\textit{you, here, tomorrow}), and that syntactic cues (e.g. quotation) contribute to LLM performance with some indexicals, while they reduce performance with others. Code and data are available at: \url{https://github.com/metehanoguzz/LLMs-Indexicals-English}

\end{abstract}

\section{Introduction}

Large Language Models (LLMs) have demonstrated remarkable capabilities in zero-shot and few-shot learning, excelling across a wide array of tasks such as machine translation, text summarization, and question answering \cite{openai2024gpt4, ye2023comprehensive, bakman2024mars, yaldiz2024designlearntrainablescoring}. Their versatility has led to widespread applications in diverse domains, including education, law, and medicine. 

\begin{figure*}[!htbp]
\vskip -0.2in
\centering
\includegraphics[width=0.95\linewidth]{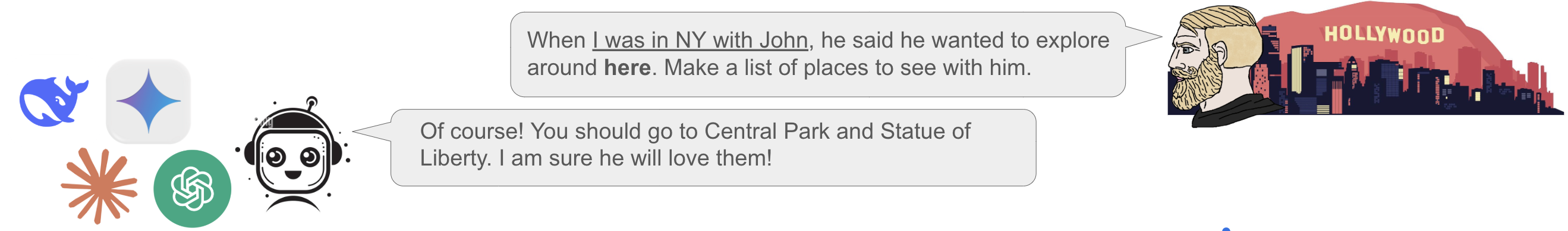}
\vskip -0.15in
\caption{An example for LLM misinterpreting indexical element `here', uttered by a speaker in Los Angeles.}
\label{fig:overview}
\end{figure*}

As the use of LLMs continues to expand, understanding their underlying behaviors has become increasingly important. Recent studies have evaluated the performance of large language models on linguistic tasks such as coreference resolution \cite{assessllmcoreference, arellmscoreference, LLMsfewshot, gpt2pronoun_res, coreference_clinical}.

Previous work on coreference resolution mostly focused on how coreference is established between two third person entities such as proper names (e.g. \textit{Andy, the mechanic}) and third person pronouns (e.g. \textit{he, him, himself}) in English and other languages \citep[e.g.][]{gpt2pronoun_res, yang-2025-language}. In this study, we investigate how LLMs establish coreference with \textit{indexical} elements (e.g. \textit{I, you, here}), which differ from third person nouns/pronouns in substantial ways and bring unique challenges for LLMs (see Figure \ref{fig:overview} as an example).
We investigate how state-of-the-art LLMs interpret the indexical elements \textit{I, you, here} and \textit{tomorrow} in English sentences, and whether context or grammatical constraints influence their decisions. To the best of our knowledge, this is the first study examining LLMs' handling of indexical elements in English. Our key contributions are as follows:
\begin{itemize}[topsep=0.7pt]
\itemsep -0.08in
    \item We introduce the \textbf{English Indexical Dataset}, comprising 400 interpretation samples for each indexical element, \textit{I, you, here, tomorrow}, totaling 1,600 instances.
    \item We evaluate the performance of four frontier LLMs, GPT-4o, Claude 3.5 Sonnet, Gemini 1.5 Pro, and DeepSeek-V3, on the interpretation of indexical elements in English.
    \item We show that LLM performances are not uniform across different types of indexical elements: indexical \textit{I} is successfully interpreted by most LLMs, other indexicals like \textit{you}, \textit{here} and \textit{tomorrow} lead to poor performances.
    \item We show that quotation affects LLMs performances differently: quotation reduces LLMs' accuracies with \textit{tomorrow}, but it increases accuracies with \textit{here}.
\end{itemize}

\section{Indexical elements} \label{indexvspron}

Indexical pronominals like \textit{I, you, here} and \textit{now} are used to refer to referents of the speech-act coordinates \citep[e.g.][]{kaplan:1977, schlenker:2003}. For instance, \textit{I} refers to \textit{author} (speaker) of the utterance, while \textit{here} refers to the \textit{location} where the utterance was made. Thus, a sentence like \ref{Eng} means different things if uttered by different people and/or in different places. If \textit{John} utters \ref{Eng1} in \textit{Los Angeles}, it means that John was born in Los Angeles, but if \textit{Mary} utters the same sentence in \textit{New York} it means that Mary was born in New York.

\vspace{-5pt}
\ex. \label{Eng}
\a. \textbf{I} was born \textbf{here}. \label{Eng1} 
\b. Andrew said that \textbf{I} went to Buckhead.\label{Eng2} 
\vspace{-17pt}

\noindent Indexicals are interpreted inside the context of utterance, referring to the actual speech-act coordinates like the \textit{author} or the \textit{location} of the actual utterance. As a result, if \ref{Eng2} is uttered by \textit{John}, the indexical \textit{I} can only be interpreted as referring to John as the speaker, leading to a reading like `Andrew said that \textbf{John} went to Buckhead'. Crucially, even though Andrew's speech/claim is reported in \ref{Eng2}, the indexical \textit{I} cannot refer to Andrew.

Direct quotation is an exception to this generalization, where reported material is interpreted as verbatim utterance/thoughts of its owner. Thus, when indexicals appear inside direct quotation, they are interpreted inside the reported context, rather than the actual context of utterance. In other words, direct quotation `shifts' the interpretations of indexicals into the reported context. For example, \textit{I} and \textit{here} in \ref{quot} appear inside direct quotation, where Andrew's speech is reported. 

\vspace{-5pt}
\ex. While we were in Atlanta, Andrew said ``\textbf{I} was born \textbf{here}.'' \label{quot}
\vspace{-5pt}

Regardless of who utters \ref{quot}, the sentence means that \textbf{Andrew} was born in \textbf{Atlanta}, so both \textit{I} and \textit{here} are `shifted' into the reported context, where Andrew is the \textit{speaker} and Atlanta is the \textit{location}\footnote{Some languages (not English) allow indexical elements to `shift' without quotation. See \citet{deal:2020} for an overview.}

Indexicals differ from other pronominals in substantial ways. First, though syntactic and semantic factors can affect how a pronoun is interpreted (e.g. subject bias), pronouns are typically ambiguous regarding what/who they refer to. For example, the third person pronoun \textit{he} in \ref{Eng3} is most naturally interpreted as referring to the subject \textit{John} (for syntactic or contextual reasons), but the object \textit{Bill} is still a possible antecedent, causing ambiguity between two different readings \cite[e.g.][]{crawley-etal:1990, stewart-pickering:1998, pickering-majid:2007}. In addition, \textit{he} can refer to any contextually salient person that is not mentioned in the sentence (e.g. \textit{Peter}), which makes pronouns even more ambiguous and context-dependent.
\vspace{-3pt}
\ex. John hit Bill and \textbf{he} ran away. \label{Eng3} 
\vspace{-3pt}

\noindent  As a result, semantic/contextual information plays a crucial role in how pronouns are interpreted, and speakers use those cues to establish coreference with pronouns. For instance, if \ref{Eng4} is uttered in a context where John is a supportive and humble coworker, the most natural interpretation is that John suggests that Bill should get promoted (\textit{he} = Bill). However, if John is arrogant and jealous, the most natural interpretation is that John suggests that John should get promoted (\textit{he} = John). 
\vspace{-3pt}
\ex. John told Bill that \textbf{he} should get promoted. \label{Eng4} 
\vspace{-3pt}

\noindent Indexicals, on the other hand, unambiguously refer to the referents of the speech-act coordinates. For example, \textit{I} in \ref{Eng5} refers to the speaker regardless of what we know about John or Bill. \textit{I} refers to the speaker even if \textit{John} and/or \textit{Bill} are arrogant and jealous, so contextual information like this should be disregarded while interpreteting indexicals.

\vspace{-3pt}
\ex. John told Bill that \textbf{I} should get a promotion.\label{Eng5} 
\vspace{-15pt}

In summary, indexicals are restricted by different syntactic factors than pronouns (e.g. quotation vs non-quotation) and are typically unambiguous, while pronouns are free to refer to a wide range of entities. Thus, indexical elements create a unique challenge for LLMs, requiring to `disregard' semantic/contextual cues that might prime interpretations through other antecedents (unlike pronouns).

\section{Experimental design}

\subsection{Dataset Curation} \label{dataset}

We curated a dataset specifically designed to test how LLMs interpret indexical elements like \textit{I, you, here,} and \textit{tomorrow} in different contexts of utterance. Specifically, we assess how these models interpret indexicals in `shifted' context prime, where context more naturally requires the indexical should be interpreted inside reported context (e.g. \textit{Peter is one of the most arrogant students in my classroom. ... Peter says that \textbf{I} am smart.}) vs 'non-shifted' context prime, where context more naturally requires the indexical should be interpreted inside actual speech context (e.g. \textit{Peter is very kind and supportive. ... Peter says that \textbf{I} am smart.}). We also included direct quotations in both contexts (e.g., \textit{Peter says, ``I am smart''}) to examine if LLMs can successfully consider syntactic factors (quotation vs regular sentences) while ignoring misleading information from the context during coreference resolution with indexical elements.

For each type of indexical, we design 100 sentences and for each sentence we apply the four different transformations explained above. Overall, we have 400 samples per indexical, compromising a total dataset of size 1600. 

We utilize GPT-4o to curate the dataset, by giving a detailed description of the task along with some in-context examples. Then it is asked to generate scenarios with a stimulus sentence in two different contexts (See Appendix~\ref{sec:dataset_gen_prompts} for the prompts used), along with specific questions addressing the referent of the indexical in each sentence. To ensure the quality of the dataset, 25\% of the dataset (400 trials = 100 sentences in four conditions) was randomly selected for evaluation, and the evaluation process consisted of three steps.

In the first step, we confirmed that all sentences were grammatically correct. In the second step, we confirmed that all quotation condition sentences had quoted embedded clauses, and all non-quotation condition sentences had regular (non-quoted) embedded clauses. We also made sure that the two sentence conditions were maximally similar, except for the quotation vs non-quotation status (i.e. the only difference between two sentence conditions was the quotation). In the third step, we checked the context prime texts for each condition in each sentence, making sure that the correct readings (shifted vs non-shifted) were primed by the context description. For example, for an item condition where \textit{here} was supposed to be shifted, we confirmed that the context description would be most naturally followed by a sentence where \textit{here} would be shifted.

100 trials from each indexical item (25\% for each indexical), 400 trials in total, were randomly selected to make sure that the evaluation/confirmation was representative of all indexical item conditions (i.e. items with \textit{I, you, here} and \textit{tomorrow}).

To eliminate potential gender bias, each dataset sample exclusively uses either male or female names, alternating to ensure a balanced distribution with 50\% of the samples containing female names and 50\% male names. This method promotes gender neutrality across the dataset. Sample details are in Appendix~\ref{sec:sample_dataset}, and the complete dataset is available in the supplementary materials.

\subsection{Models} In our evaluation, we utilize four recent state-of-the-art LLMs: GPT-4o \cite{openai2024gpt4},  Claude 3.5 Sonnet \cite{TheC3}, Gemini 1.5 Pro \cite{geminiteam2024gemini15unlockingmultimodal}, and DeepSeek-V3 \cite{deepseekai2024deepseekv3technicalreport}. This selection of diverse models provides a comprehensive evaluation of LLM performance with indexicals.

\subsection{Evaluation Strategy} To assess the performance of the model, we specifically prompt it to answer questions designed to test its capabilities as described in Section \ref{dataset}. Additionally, to ensure focused responses, we restrict the model's answers to one of two predefined options: the `shifted' option and the `non-shifted' option. We provide the prompt in Appendix~\ref{sec:eval_prompt}.

\subsection{Metrics.} 
We assess model accuracy across four cases for each indexical: (i) Non-quoted sentences with shifted context prime, (ii) Non-quoted with non-shifted prime, (iii) Quoted with shifted prime, and (iv) Quoted with non-shifted prime. Optimal performance would be achieved by always selecting the `shifted' option in quoted conditions and selecting the `non-shifted' option in non-quoted conditions.

\begin{figure*}
\begin{center}
\includegraphics[width=0.24\textwidth]{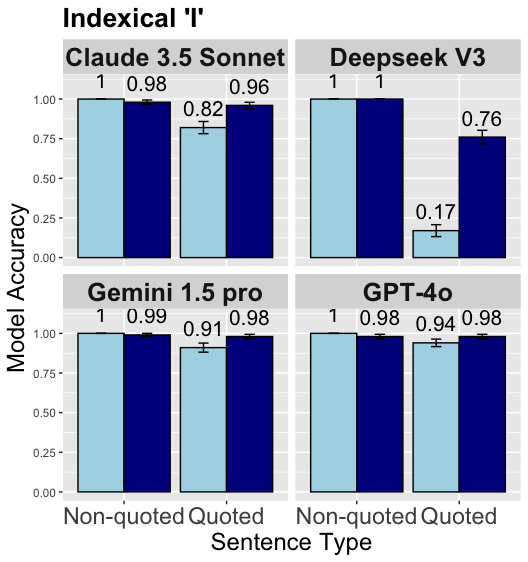}
\includegraphics[width=0.24\textwidth]{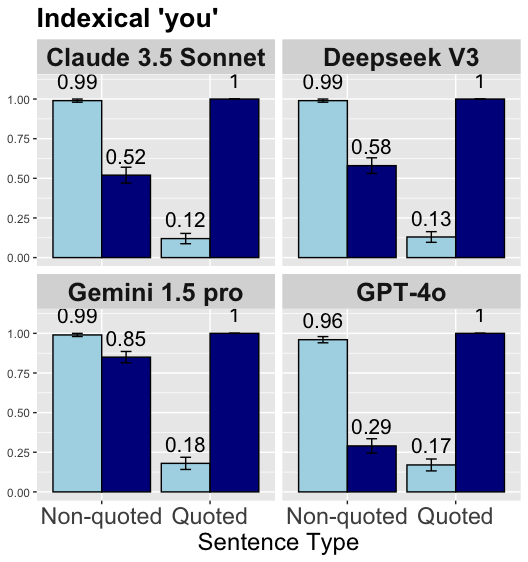}
\includegraphics[width=0.24\textwidth]{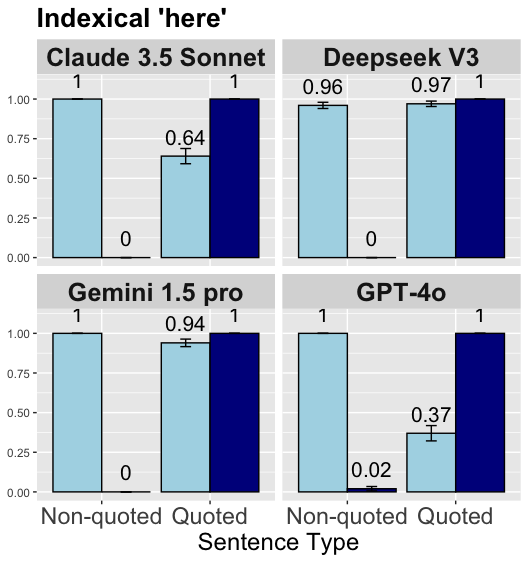}
\includegraphics[width=0.24\textwidth]{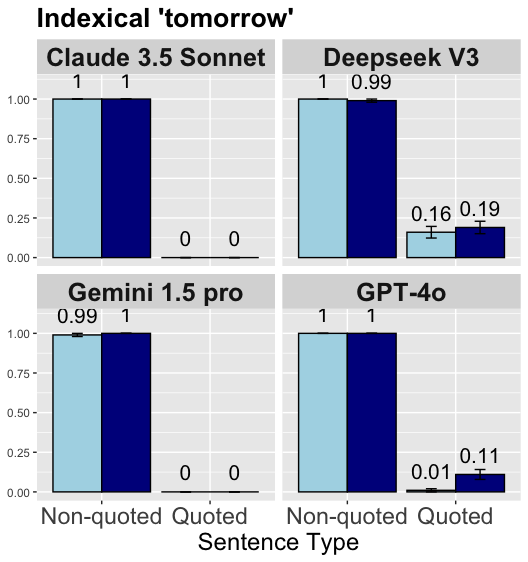}
\vskip -0.1in
\caption{From left to right: Performance analysis plot of for the indexical `I', Performance analysis plot of for the indexical `you', Performance analysis plot of for the indexical `here', Performance analysis plot of for the indexical `tomorrow'. Dark blue bars = Shifted context prime, Light blue bars = Non-shifted context prime.}
\vskip -0.3in
\label{results}
\end{center}
\end{figure*}

\section{Experimental Results}

We present the experimental results in Figure~\ref{results} and we discuss them in detail for each indexical type: 

\paragraph{Indexical \textit{I}.} The results show that all four models perform near optimum in sentences without quotation with an average accuracy of 99\% (always correctly selecting the non-shifted option). For the quoted sentences GPT-4o and Gemini 1.5 pro again perform almost optimum with a mean accuracy larger than 94\%, followed by Claude 3.5 Sonnet with 89\% mean accuracy (correctly selecting the shifted option). However, DeepSeek V3 fails to always select the shifted option on quoted sentences. Especially, the model performance drops significantly to 17\% when context primes non-shifted readings, suggesting that inclusion of quotation makes the model more sensitive to the linguistically irrelevant effects of context prime. Moreover, considering that the model performance only reaches 78\% accuracy in quotation conditions seems to imply that the model might have a bias towards the non-shifted reading overall, reducing model performance in quotation conditions.

\paragraph{Indexical \textit{you}.} The results indicate that LLMs generally perform worse with the indexical \textit{you} than with \textit{I}. We see similar patterns across models, suggesting they perform mostly at similar levels. All models are sensitive to the effects of context prime, performing lower when the context primes the incorrect option (i.e. shifted reading in non-quotation and non-shifted reading in quotation). Moreover, quotation consistently results in lower performance across models. 
Notably, Gemini 1.5 Pro excels in non-quotation accuracy (92\%), though its performance drops significantly under quotation conditions to the levels of other models. 
Overall, the results suggest that models interpret \textit{you} based on the linguistically irrelevant context prime rather than the sentence type, which indeed determines the correct readings of the indexicals.

\paragraph{Indexical \textit{here}.} The results show that LLMs, similar to \textit{you}, struggle to interpret the indexical \textit{here}, especially when context primes the incorrect option, leading to poor performance.
However, different from \textit{you}, quotation leads to higher performance with \textit{here}. In non-quotation conditions, all LLMs make their selections almost exclusively based on context prime, with shifted primes showing over 96\% accuracy and non-shifted primes less than 2\%, 
where context prime ideally should not have any effects on the selection. In contrast, under quotation conditions, the influence of context prime diminishes, leading to higher performances. Particularly, DeepSeek V3 and Gemini 1.5 Pro exhibit impressive performance with accuracies above 97\% and 94\%, respectively, followed by Gemini 1.5 Pro and GPT-4o, with accuracies above 64\% and 37\%.

\paragraph{Indexical \textit{tomorrow}.} The results indicate that LLMs have a strong bias towards the non-shifted interpretations of \textit{tomorrow}. We see that the models almost always select the non-shifted option for trials with \textit{tomorrow}, regardless of context prime or sentence type. While Claude 3.5 Sonnet and Gemini 1.5 Pro selects the non-shifted options 100\% of the time, GPT-4o and DeepSeek V3 select the non-shifted option 94\% and 83\% of the time, respectively. This strong bias leads to illusory high accuracies in non-quotation conditions, while causes extremely low performance in quotation conditions.

\section{Related Work}
Coreference resolution has been extensively studied in prior research \cite{assessllmcoreference, arellmscoreference, LLMsfewshot, gpt2pronoun_res, coreference_clinical}. However, indexical elements exhibit distinct syntactic properties compared to other (non-indexical) pronominals, as discussed in Section \ref{indexvspron}. Previous work by \citet{oguz-etal-2024-llms} explored how LLMs interpret indexicals in Turkish, where indexicals possess different grammatical properties than in English and can shift without quotation. To the best of our knowledge, this study is the first to evaluate the performance of LLMs in interpreting indexicals in English.

\section{Discussion and Conclusion}

Our results show that LLM performances are not uniform across different types of indexical elements and sentence types. While most LLMs perform at impressive levels interpreting the indexical \textit{I}, their performances drop significantly in other indexicals, particularly in \textit{here} and \textit{tomorrow}. Moreover, though sentence type (quotation vs non-quotation) affects LLMs performances in general, the effects are not similar across different indexical types and models. While quotation increases LLMs' performance with \textit{here}, it decreases their performance with \textit{you} and \textit{tomorrow}. In addition, \textit{tomorrow} seems to be affected by quotation in a greater magnitude than \textit{you}. In conclusion, we find that different types of indexicals show unique patterns regarding how they are interpreted by the LLMs.

Our results diverge from those reported in \citet{oguz-etal-2024-llms}, who tested how first person indexical in Turkish \textit{ben} `I' was interpreted by different LLMs, including GPT-4o and show that LLMs exhibit very poor performance interpreting \textit{ben} `I'. Here, we report that LLMs perform at an almost human-like level with interpreting the English indexical \textit{I}. This might be due to lower amounts of available resources to train the models in Turkish, compared to English. Another reason for low performance in Turkish could be due to the fact that Turkish is a \textit{pro}-drop language, meaning that the subject pronouns can be dropped (silent). \citet{oguz-etal-2024-llms} used sentences where the first person indexical \textit{ben} `I' was dropped, which might have made the task more challenging for the LLMs considering that dropped indexicals can show different properties than overt ones \cite{oguz-etal:2020}. This linguistic difference between Turkish and English might have caused different results between the Turkish and English tests.

\section{Limitations}
Our work explores how LLMs interpret indexical elements in a black-box setting but does not provide experimental analyses that investigate the underlying reasons for these behaviors by examining the models' internals or training data. Future research could adopt a white-box approach to analyze these behaviors in greater depth, offering valuable insights into the mechanisms driving LLMs' interpretation of indexicals.

\bibliography{main}

\newpage

\appendix

\section{Dataset Details}
\label{sec:dataset details}

\subsection{Dataset Generation Prompts}\label{sec:dataset_gen_prompts}

We provide the prompts used for data generation in Tables \ref{tab:you_prompt}, \ref{tab:tmr+prompt}, \ref{tab:I_prompt}, and \ref{tab:here_prompt}. The prompts used for question generation per scenario is presented in Tables \ref{tab:I_qprompt}, \ref{tab:you_qprompt}, \ref{tab:here_qprompt}, and \ref{tab:tmr_qprompt}.

\begin{table*}[ht]
\centering
\begin{tabular}{|p{\textwidth}|}
\hline

I would like you to help me create a stimulus for my project. The stimuli will be English sentences. For each sentence, there will be two types of context description. One context description will prime the actual meaning of the sentence, but the other description will prime an incorrect reading of the sentence (like a misinterpretation). I want you to follow a structure while creating sentences and contexts. You can find more details below: \\ \\ 

Details for stimulus sentences: Each stimulus sentence will have a structure like "While John was speaking to Travis, he said that Chris appreciates you a lot." Please make sure to use the names John, Travis, and Chris. Make sure that Chris is the subject of the embedded clause, and make sure that the sentence begins as “While John was speaking to Travis…” Please use different embedded verbs. For example, rather than appreciates you a lot, you can use saw you at the market, etc. But the action should be done by Chris, and the object should be "you". \\ \\ 

Details about the context descriptions: The contexts will prime how "you" in the stimulus is interpreted. To manipulate this, I will give contexts in which "you" would refer to my addressee, which is you. But, to prime the incorrect interpretation, I will give contexts that would naturally follow if "you" referred to John’s addressee. However, since “you” means the current addressee, this will be a misinterpretation. For example, a correct reading prime would be like "Hi! I am Andrew. I will call you Donald. I am a graduate student and I live in Los Angeles. I have two friends named John and Chris. Chris asked for help from you a few times in the past, and you always helped him." This context makes it sound like Chris would appreciate you for all your help, and thus it would not be surprising if Chris appreciated you (Donald). However, in the incorrect interpretation prime, I will use a context like "Hi! I am Andrew. I will call you Donald. I am a graduate student and I live in Los Angeles. I have two friends named John and Chris. Chris asked for help from Travis a few times in the past, and Travis always helped him." In this context, it would be more natural if Chris appreciated Travis, who helped him, instead of you (Donald), and thus would be natural if "you" in the sentence was interpreted as referring to Travis (though “you” should refer to my addressee, which is you (Donald)). This would create the incorrect misinterpretation. Please make sure to keep the person names constant. Also, make sure that the context starts exactly as “Hi! I am Andrew. I will call you Donald. I am a graduate student and I live in Los Angeles.” \\ \\ 

These are some examples you generated before: \{previous\_generations\}. Use these examples as inspiration to spark creativity. Provide one new stimulus sentence, along with a corresponding correct prime context and an incorrect prime context, in the following format:  

stimulus\_sentence:  stimulus stimulus sentence

correct\_context: correct prime context sentence 

wrong\_context: incorrect prime context sentence \\

\hline
\end{tabular}
\caption{Prompt used to generate dataset samples for indexical `you'.}
\label{tab:you_prompt}
\end{table*}

\begin{table*}[ht]
\centering
\begin{tabular}{|p{\textwidth}|}
\hline

I would like you to help me create a stimulus for my project. The stimuli will be English sentences. For each sentence, there will be two types of context description. One context description will prime the actual meaning of the sentence, but the other description will prime an incorrect reading of the sentence (like a misinterpretation). I want you to follow a structure while creating sentences and contexts. You can find more details below: \\ \\

Details for stimulus sentences: Each stimulus sentence will have a structure like "When we spoke last summer, John said that Chris was going to go to Greece tomorrow." Please make sure to use the names John, and Chris. Make sure that Chris is the subject of the embedded clause. Please use different embedded verbs. For instance, rather than go to Greece, you can use have a visa appointment, etc. But the action should be done "tomorrow". \\ \\

Details about the context descriptions: The contexts will prime how "tomorrow" in the stimulus is interpreted. To manipulate this, I will give contexts in which "tomorrow" would refer to the day after the actual (matrix) sentence is uttered. But, to prime the incorrect interpretation, I will give contexts that would naturally follow if "tomorrow" referred to the day after John spoke. However, since “tomorrow” means the day after the current day, this will be a misinterpretation. For example, a correct reading prime would be like "Hi! I am Andrew. I am a graduate student and I live in Los Angeles. I have two friends named John and Chris. Chris is very careful about planning everything, and always plans his stuff ahead of time." This context makes it sound like Chris would plan his trip to Greece ahead of time, and thus it would not be surprising if John said last summer that Chris was going to have a trip tomorrow. However, in the incorrect interpretation prime, I will use a context like "Hi! I am Andrew. I am a graduate student and I live in Los Angeles. I have two friends named John and Chris. Chris is very lazy and never plans his stuff until the last moment." In this context, it would be more natural if Chris was going to go to Greece last summer, the day after John spoke to me, instead of the day after today, and thus would be natural if "tomorrow" in the sentence was interpreted as the day after John spoke to me (though “tomorrow” should refer to the day after today). This would create the incorrect misinterpretation. Please make sure to keep the person names constant. Also, make sure that the context starts exactly as “Hi! I am Andrew. I am a graduate student and I live in Los Angeles.” \\ \\

These are some examples you generated before: \{previous\_generations\}. Use these examples as inspiration to spark creativity. Provide one new stimulus sentence, along with a corresponding correct prime context and an incorrect prime context, in the following format:  

stimulus\_sentence:  stimulus stimulus sentence

correct\_context: correct prime context sentence 

wrong\_context: incorrect prime context sentence \\

\hline
\end{tabular}
\caption{Prompt used to generate dataset samples for indexical `tomorrow'.}
\label{tab:tmr+prompt}
\end{table*}

\begin{table*}[ht]
\centering
\begin{tabular}{|p{\textwidth}|}
\hline

I would like you to help me create a stimulus for my project. The stimuli will be English sentences. For each sentence, there will be two types of context description. One context description will prime the actual meaning of the sentence, but the other description will prime an incorrect reading of the sentence (like a misinterpretation). I want you to follow a structure while creating sentences and contexts. You can find more details below: \\ \\ 

Details for stimulus sentences: Each stimulus sentence will have a structure like "Chris thinks that I will win the race." Please make sure to use the name Chris as the matrix subject, and “I” as the embedded subject subject. Please use different embedded verbs. For example, rather than win the race, you can use study hard, etc. But the action should be done by "I". \\ \\

Details about the context descriptions: The contexts will prime who "I" in the stimulus refers to. To manipulate this, I will give contexts in which "I" would refer to the speaker. But, to prime the incorrect interpretation, I will give contexts that would naturally follow if "I" referred to Chris. However, since Chris is not the speaker, this will lead to an incorrect interpretation. 

For example, a correct reading prime would be like "Hi! I am Andrew. I am a graduate student and I live in Los Angeles. I have a friend named Chris. Chris is a supportive friend, and has always trusted in my abilities. There is a race next week." This context makes it sound like Chris would support the speaker in a race, and predict that the speaker would win the race. However, in the incorrect interpretation prime, I will use a context like "Hi! I am Andrew. I am a graduate student and I live in Los Angeles. I have a friend named Chris. Chris is usually very competitive, and has bullied other people in front of me. There is a race next week.." In this context, it would be more natural if Chris thought that he would win the race, instead of the speaker Andrew, and thus would be natural if "I" in the sentence referred to Chris (though “I” should refer to the speaker Andrew). This would create the incorrect misinterpretation. Please make sure to keep the person names constant. Also, make sure that the context starts exactly as “Hi! I am Andrew. I am a graduate student and I live in Los Angeles." \\ \\

These are some examples you generated before: \{previous\_generations\}. Use these examples as inspiration to spark creativity. Provide one new stimulus sentence, along with a corresponding correct prime context and an incorrect prime context, in the following format:  

stimulus\_sentence:  stimulus stimulus sentence

correct\_context: correct prime context sentence 

wrong\_context: incorrect prime context sentence
 \\

\hline
\end{tabular}
\caption{Prompt used to generate dataset samples for indexical `I'.}
\label{tab:I_prompt}
\end{table*}

\begin{table*}[ht]
\centering
\begin{tabular}{|p{\textwidth}|}
\hline

I would like you to help me create a stimulus for my project. The stimuli will be English sentences. For each sentence, there will be two types of context description. One context description will prime the actual meaning of the sentence, but the other description will prime an incorrect reading of the sentence (like a misinterpretation). I want you to follow a structure while creating sentences and contexts. You can find more details below: \\ \\

Details for stimulus sentences: Each stimulus sentence will have a structure like "When I was in New York with John, he said that Chris wanted to explore here." Please make sure to use the city name New York for the sentence, and the names John and Chris. John will always be the person who says something that Chris will do. In each sentence, Chris will be the person who is doing something "here". Please use different verbs. For example, rather than explore here, you can use attend a conference here, etc.. Use various verbs. But the action should be done "here". \\ \\

Details about the context descriptions: The contexts will prime where "here" in the stimulus refers to. To manipulate this, I will give contexts in which "here" would refer to Los Angeles, where the author/speaker of the sentence is located. But, to prime the incorrect interpretation, I will give contexts that would naturally follow if "here" referred to New York. However, since the speaker is not in New York, this will lead to an incorrect interpretation. 

For example, a correct reading prime would be like "Hi! I am Andrew. I am a graduate student and I am studying in a cafe in Los Angeles. I have two friends named John and Chris. They love exploring new cities, and surprisingly they did not spend much time in Los Angeles." This context makes it sound like John and Chris would like to come to Los Angeles to explore around, and thus the "explore around here" in the stimulus sentence would be naturally understood as Los Angeles. However, in the incorrect interpretation prime, I will use a context like "Hi! I am Andrew. I am a graduate student and I am studying in a cafe in Los Angeles. I have two friends named John and Chris. They love exploring new cities, and surprisingly they did not spend much time in New York." In this context, it would be more natural if John and Chris would like to go to New York, and thus would be natural if "explore around here" would be exploring around New York (though here should refer to the place where the speaker Andrew is, which is Los Angeles). This would create the incorrect misinterpretation. Please make sure to keep the person names and city names constant. 

Also, make sure that the context starts exactly as “Hi! I am Andrew. I am a graduate student and I am studying in a cafe in Los Angeles.”. \\ \\

These are some examples you generated before: \{previous\_generations\}. Use these examples as inspiration to spark creativity. Provide one new stimulus sentence, along with a corresponding correct prime context and an incorrect prime context, in the following format:  

stimulus\_sentence:  stimulus stimulus sentence

correct\_context: correct prime context sentence 

wrong\_context: incorrect prime context sentence\\

\hline
\end{tabular}
\caption{Prompt used to generate dataset samples for indexical `here'.}
\label{tab:here_prompt}
\end{table*}
\begin{table*}[ht]
\centering
\begin{tabular}{|p{\textwidth}|}
\hline

I would like you to help me create questions to the sentences I will give you for my project. The question will be English sentences. The goal is to create a question that will prime the correct interpretation of the indexical in the sentence. \\ \\

Here are examples: \\ \\

Sentences:

Hi! I am Andrew. I am a graduate student and I live in Los Angeles. I have a friend named Chris. Chris knows how dedicated I am to my studies and always encourages me to do my best. The final exams are coming up next week. Chris thinks that I will get good grades on the final exams.

Hi! I am Andrew. I am a graduate student and I live in Los Angeles. I have a friend named Chris. Chris usually outperforms everyone in our classes and often brags about it. The final exams are coming up next week. Chris thinks that I will get good grades on the final exams. \\ \\

Question: According to the last sencence, who will get good grades on the final exams?  \\ \\

Sentences:

Hi! I am Andrew. I am a graduate student and I live in Los Angeles. I have a friend named Chris. Chris believes in my ability to communicate effectively and often praises my public speaking skills. There is an important conference coming up soon. Chris thinks that I will give a great presentation at the conference.

Hi! I am Andrew. I am a graduate student and I live in Los Angeles. I have a friend named Chris. Chris is known for his exceptional presentation skills and often receives awards for them. There is an important conference coming up soon. Chris thinks that I will give a great presentation at the conference.  \\ \\

Question: According to the last sencence, who will give a great presentation at the conference?  \\ \\

Now, for the following sentences, you should create a question whose answer can only be either Chris or Andrew. The question should be designed to prime the correct interpretation of the indexical in the sentence. Provide only the question and always start with "According to the last sentence, who"  \\ \\

Sentences:

\{sentences\} \\ \\

Question: \\

\hline
\end{tabular}
\caption{Prompt used to generate question for samples containing indexical `I'.}
\label{tab:I_qprompt}
\end{table*}

\begin{table*}[ht]
\centering
\begin{tabular}{|p{\textwidth}|}
\hline
I would like you to help me create questions to the sentences I will give you for my project. The question will be English sentences. The goal is to create a question that will prime the correct interpretation of the indexical in the sentence. \\ \\

Here are examples:\\ \\

Sentences:

Hi! I am Andrew. I will call you Donald. I am a graduate student and I live in Los Angeles. I have two friends named John and Chris. You and Chris attended the same conference last month, and you even had a conversation there. While John was speaking to Travis, he said that Chris saw you at the conference.

Hi! I am Andrew. I will call you Donald. I am a graduate student and I live in Los Angeles. I have two friends named John and Chris. Travis and Chris attended the same conference last month, and they even had a conversation there. While John was speaking to Travis, he said that Chris saw you at the conference.\\ \\

Question: According to the last sentence, who did John say that Chris saw at the conference?\\ \\

Sentences:

Hi! I am Andrew. I will call you Donald. I am a graduate student and I live in Los Angeles. I have two friends named John and Chris. You and Chris are both art enthusiasts and attended the same art gallery event last weekend, where you ran into each other. While John was speaking to Travis, he said that Chris noticed you at the art gallery.

Hi! I am Andrew. I will call you Donald. I am a graduate student and I live in Los Angeles. I have two friends named John and Chris. Travis and Chris are both art enthusiasts and attended the same art gallery event last weekend, where they ran into each other. While John was speaking to Travis, he said that Chris noticed you at the art gallery.\\ \\

Question: According to the last sentence, who did John say that Chris noticed at the art gallery?\\ \\

Now, for the following sentences, you should create a question whose answer can only be either Travis or Donald. The question should be designed to prime the correct interpretation of the indexical in the sentence. Provide only the question and always start with "According to the last sentence, who did John say that Chris"\\ \\

Sentences:

\{sentences\} \\ \\

Question: \\

\hline
\end{tabular}
\caption{Prompt used to generate question for samples containing indexical `you'.}
\label{tab:you_qprompt}
\end{table*}

\begin{table*}[ht]
\centering
\begin{tabular}{|p{\textwidth}|}
\hline
I would like you to help me create questions to the sentences I will give you for my project. The question will be English sentences. The goal is to create a question that will prime the correct interpretation of the indexical in the sentence.\\ \\

Here are examples:\\ \\

Sentences:

Hi! I am Andrew. I am a graduate student and I am studying in a cafe in Los Angeles. I have two friends named John and Chris. They are both academics who love to participate in international conferences. Recently, I've been telling them about the exciting academic events happening right here in Los Angeles. When I was in New York with John, he said that Chris wanted to attend a conference here.

Hi! I am Andrew. I am a graduate student and I am studying in a cafe in Los Angeles. I have two friends named John and Chris. They are both academics who love to participate in international conferences. Recently, they realized they haven't attended many conferences in New York, which is quite surprising given their love for the city. When I was in New York with John, he said that Chris wanted to attend a conference here.\\ \\

Question: According to the last sentence, where does Chris want to attend a conference?\\ \\

Sentences:

Hi! I am Andrew. I am a graduate student and I am studying in a cafe in Los Angeles. I have two friends named John and Chris. They both have a deep appreciation for art, and recently I've been telling them about the vibrant art scene here in Los Angeles that offers great opportunities for new gallery openings. When I was in New York with John, he said that Chris wanted to open an art gallery here.

Hi! I am Andrew. I am a graduate student and I am studying in a cafe in Los Angeles. I have two friends named John and Chris. They both have a deep appreciation for art, and recently they realized they haven't opened a gallery in New York yet, despite its renowned art scene, which is surprising given their passion. When I was in New York with John, he said that Chris wanted to open an art gallery here.\\ \\

Question: According to the last sentence, where does Chris want to open an art gallery?\\ \\

Now, for the following sentences, you should create a question whose answer can only be either Los Angeles or New York. The question should be designed to prime the correct interpretation of the indexical in the sentence. Provide only the question and always start with "According to the last sentence, where does Chris want to".\\ \\

Sentences:

\{sentences\} \\ \\

Question: \\

\hline
\end{tabular}
\caption{Prompt used to generate question for samples containing indexical `here'.}
\label{tab:here_qprompt}
\end{table*}

\begin{table*}[ht]
\centering
\begin{tabular}{|p{\textwidth}|}
\hline
I would like you to help me create questions to the sentences I will give you for my project. The question will be English sentences. The goal is to create a question that will prime the correct interpretation of the indexical in the sentence.\\ \\

Here are examples:\\ \\

Sentences:

Hi! I am Andrew. I am a graduate student and I live in Los Angeles. I have two friends named John and Chris. Chris is meticulous about arranging all his necessary documentation well in advance and likes having everything sorted before deadlines approach. When we spoke last summer, John mentioned that Chris would be having his visa appointment tomorrow.

Hi! I am Andrew. I am a graduate student and I live in Los Angeles. I have two friends named John and Chris. Chris is quite disorganized and often waits until the last possible moment to schedule important tasks such as visa appointments. When we spoke last summer, John mentioned that Chris would be having his visa appointment tomorrow.\\ \\
 
Question: According to the last sentence, did Chris already have his visa appointment, or is he going to do so in the future?\\ \\

Sentences:

Hi! I am Andrew. I am a graduate student and I live in Los Angeles. I have two friends named John and Chris. Chris is very disciplined and likes to plan his activities ahead of time. When we spoke last summer, John told me that Chris was planning go to the bungee jumping event tomorrow.

Hi! I am Andrew. I am a graduate student and I live in Los Angeles. I have two friends named John and Chris. Chris likes trying different activities but he is very bad at planning and he usually plans his stuff at the very last moment. When we spoke last summer, John told me that Chris was planning go to the bungee jumping event tomorrow. \\ \\

Question: According to the last sentence, did Chris already go to the bungee jumping event, or is he going to do so in the future?\\ \\

Now, for the following sentences, you should create a question in the given format. The question should be designed to prime the correct interpretation of the indexical in the sentence. Provide only the question and always structure the sentence as "According to the last sentence, did Chris already ... or is he going to do so in the future?". \\ \\

Sentences:

\{sentences\} \\ \\

Question: \\

\hline
\end{tabular}
\caption{Prompt used to generate question for samples containing indexical `tomorrow'.}
\label{tab:tmr_qprompt}
\end{table*}

\subsection{Samples From the Dataset}\label{sec:sample_dataset}

We provide samples from the dataset for each indexical element we investigate in Tables \ref{tab:dataset_samples_I}, \ref{tab:dataset_samples_you}, \ref{tab:dataset_samples_here}, and \ref{tab:dataset_samples_tomorrow}.

\begin{table*}[ht]
\centering
\begin{tabular}{|p{0.5\textwidth}|l|c|c|}
\hline

\textbf{Context+Stimuli and Question }& \textbf{\thead{Shifted\\Option}} & \textbf{\thead{Non-shifted\\Option}} & \textbf{\thead{Ground\\Truth}}\\
\hline

Hi! I am Stephen. I am a graduate student and I live in Los Angeles. I have a friend named Adam. Adam knows how dedicated I am to my studies and always encourages me to do my best. The final exams are coming up next week. Adam thinks that I will get good grades on the final exams.

\textbf{Question:} According to the last sentence, who will get good grades on the final exams? &\	Adam & Stephen & Non-shifted\\
\hline

Hi! I am Stephen. I am a graduate student and I live in Los Angeles. I have a friend named Adam. Adam usually outperforms everyone in our classes and often brags about it. The final exams are coming up next week. Adam says that I will get good grades on the final exams.

\textbf{Question:} According to the last sentence, who will get good grades on the final exams? &\	Adam & Stephen & Non-shifted\\

\hline

Hi! I am Stephen. I am a graduate student and I live in Los Angeles. I have a friend named Adam. Adam knows how dedicated I am to my studies and always encourages me to do my best. The final exams are coming up next week. Adam says "I will get good grades on the final exams".

\textbf{Question:} According to the last sentence, who will get good grades on the final exams? &\	Adam & Stephen & Shifted\\

\hline

Hi! I am Stephen. I am a graduate student and I live in Los Angeles. I have a friend named Adam. Adam usually outperforms everyone in our classes and often brags about it. The final exams are coming up next week. Adam says "I will get good grades on the final exams".

\textbf{Question:} According to the last sentence, who will get good grades on the final exams? &\	Adam & Stephen & Shifted\\

\hline

\end{tabular}
\caption{Dataset Samples for Indexical "I"}
\label{tab:dataset_samples_I}
\end{table*}

\begin{table*}[ht]
\centering
\begin{tabular}{|p{0.5\textwidth}|l|c|c|}
\hline

\textbf{Context+Stimuli and Question }& \textbf{\thead{Shifted\\Option}} & \textbf{\thead{Non-shifted\\Option}} & \textbf{\thead{Ground\\Truth}}\\
\hline

Hi! I am Jerry. I will call you Ryan. I am a graduate student and I live in Los Angeles. I have two friends named Dylan and Gregory. You and Gregory attended the same conference last month, and you even had a conversation there. While Dylan was speaking to Samuel, he said that Gregory saw you at the conference.

\textbf{Question:} According to the last sentence, who did Dylan say that Gregory saw at the conference? & Ryan & Samuel & Non-shifted\\
\hline

Hi! I am Jerry. I will call you Ryan. I am a graduate student and I live in Los Angeles. I have two friends named Dylan and Gregory. Samuel and Gregory attended the same conference last month, and they even had a conversation there. While Dylan was speaking to Samuel, he said that Gregory saw you at the conference.

\textbf{Question:} According to the last sentence, who did Dylan say that Gregory saw at the conference? & Ryan & Samuel & Non-shifted\\
\hline

Hi! I am Jerry. I will call you Ryan. I am a graduate student and I live in Los Angeles. I have two friends named Dylan and Gregory. You and Gregory attended the same conference last month, and you even had a conversation there. While Dylan was speaking to Samuel, he said "Gregory saw you at the conference".

\textbf{Question:} According to the last sentence, who did Dylan say that Gregory saw at the conference? & Ryan & Samuel & Shifted\\
\hline

Hi! I am Jerry. I will call you Ryan. I am a graduate student and I live in Los Angeles. I have two friends named Dylan and Gregory. Samuel and Gregory attended the same conference last month, and they even had a conversation there. While Dylan was speaking to Samuel, he said "Gregory saw you at the conference".

\textbf{Question:} According to the last sentence, who did Dylan say that Gregory saw at the conference? & Ryan & Samuel & Shifted\\
\hline

\end{tabular}
\caption{Dataset Samples for Indexical "You"}
\label{tab:dataset_samples_you}
\end{table*}

\begin{table*}[ht]
\centering
\begin{tabular}{|p{0.5\textwidth}|l|c|c|}
\hline

\textbf{Context+Stimuli and Question }& \textbf{\thead{Shifted\\Option}} & \textbf{\thead{Non-shifted\\Option}} & \textbf{\thead{Ground\\Truth}}\\
\hline

Hi! I am Lisa. I am a graduate student and I am studying in a cafe in Denver. I have two friends named Deborah and Jennifer. They are both academics who love to participate in international conferences. Recently, I've been telling them about the exciting academic events happening right here in Denver. When I was in Milwaukee with Deborah, she said that Jennifer wanted to attend a conference here.

\textbf{Question:} According to the last sentence, where does Jennifer want to attend a conference? & Milwaukee & Denver & Non-shifted\\
\hline

Hi! I am Lisa. I am a graduate student and I am studying in a cafe in Denver. I have two friends named Deborah and Jennifer. They are both academics who love to participate in international conferences. Recently, they realized they haven't attended many conferences in Milwaukee, which is quite surprising given their love for the city. When I was in Milwaukee with Deborah, she said that Jennifer wanted to attend a conference here.

\textbf{Question:} According to the last sentence, where does Jennifer want to attend a conference? & Milwaukee & Denver & Non-shifted\\
\hline

Hi! I am Lisa. I am a graduate student and I am studying in a cafe in Denver. I have two friends named Deborah and Jennifer. They are both academics who love to participate in international conferences. Recently, I've been telling them about the exciting academic events happening right here in Denver. When I was in Milwaukee with Deborah, she said "Jennifer wants to attend a conference here".

\textbf{Question:} According to the last sentence, where does Jennifer want to attend a conference? & Milwaukee & Denver & Shifted\\
\hline

Hi! I am Lisa. I am a graduate student and I am studying in a cafe in Denver. I have two friends named Deborah and Jennifer. They are both academics who love to participate in international conferences. Recently, they realized they haven't attended many conferences in Milwaukee, which is quite surprising given their love for the city. When I was in Milwaukee with Deborah, she said "Jennifer wants to attend a conference here".

\textbf{Question:} According to the last sentence, where does Jennifer want to attend a conference? & Milwaukee & Denver & Shifted\\
\hline

\end{tabular}
\caption{Dataset Samples for Indexical "Here"}
\label{tab:dataset_samples_here}
\end{table*}

\begin{table*}[ht]
\centering
\begin{tabular}{|p{0.5\textwidth}|l|c|c|}
\hline

\textbf{Context+Stimuli and Question }& \textbf{\thead{Shifted\\Option}} & \textbf{\thead{Non-shifted\\Option}} & \textbf{\thead{Ground\\Truth}}\\
\hline

Hi! I am Albert. I am a graduate student and I live in Los Angeles. I have two friends named Donald and Carl. Carl is meticulous about arranging all his necessary documentation well in advance and likes having everything sorted before deadlines approach. When we spoke last summer, Donald mentioned that Carl would be having his visa appointment tomorrow.

\textbf{Question:} According to the last sentence, did Carl already have his visa appointment, or is he going to do so in the future? & will do it in the future & did it in the past & Non-shifted\\
\hline

Hi! I am Albert. I am a graduate student and I live in Los Angeles. I have two friends named Donald and Carl. Carl is quite disorganized and often waits until the last possible moment to schedule important tasks such as visa appointments. When we spoke last summer, Donald mentioned that Carl would be having his visa appointment tomorrow.

\textbf{Question:} According to the last sentence, did Carl already have his visa appointment, or is he going to do so in the future? & will do it in the future & did it in the past & Non-shifted\\
\hline

Hi! I am Albert. I am a graduate student and I live in Los Angeles. I have two friends named Donald and Carl. Carl is meticulous about arranging all his necessary documentation well in advance and likes having everything sorted before deadlines approach. When we spoke last summer, Donald said "Carl will be having his visa appointment tomorrow".

\textbf{Question:} According to the last sentence, did Carl already have his visa appointment, or is he going to do so in the future? & will do it in the future & did it in the past & Shifted\\
\hline

Hi! I am Albert. I am a graduate student and I live in Los Angeles. I have two friends named Donald and Carl. Carl is quite disorganized and often waits until the last possible moment to schedule important tasks such as visa appointments. When we spoke last summer, Donald said "Carl will be having his visa appointment tomorrow".

\textbf{Question:} According to the last sentence, did Carl already have his visa appointment, or is he going to do so in the future? & will do it in the future & did it in the past & Shifted\\
\hline

\end{tabular}
\caption{Dataset Samples for Indexical "Tomorrow"}
\label{tab:dataset_samples_tomorrow}
\end{table*}

\section{Experimental Details}

\subsection{Prompt Used in Evaluation}\label{sec:eval_prompt} 

The prompt we employ for the evaluation is as follows:
\begin{verbatim}
Read the following passage carefully 
and answer the question at the end:
{stimuli}
{question}
Please provide your answer as: either
{option1} or {option2}. Do not include
any additional explanation or text.
\end{verbatim}

\end{document}